\documentclass{article}

\usepackage[preprint,nonatbib]{neurips_2022}

\usepackage[utf8]{inputenc} 
\usepackage[T1]{fontenc}    
\usepackage{hyperref}       
\usepackage{url}            
\usepackage{booktabs}       
\usepackage{amsfonts}       
\usepackage{nicefrac}       
\usepackage{microtype}      
\usepackage{xcolor}         
\usepackage{colortbl}
\usepackage{enumitem}
\usepackage{color}
\usepackage{bm} 
\usepackage{multirow}
\usepackage{array}
\usepackage{amsfonts}
\usepackage{amsmath}
\usepackage{amsthm}
\usepackage{comment,soul}
\usepackage{subcaption}
\usepackage{xspace}
\usepackage{wrapfig}
\usepackage{graphicx}
\usepackage[linesnumbered,boxed,ruled,vlined]{algorithm2e}
\usepackage{listings}

\newtheorem{definition}{Definition}

\usepackage{diagbox}
\usepackage{pifont}
\usepackage{arydshln}
\usepackage{booktabs}
\usepackage{float}
\usepackage[title]{appendix}
\usepackage{fontawesome5}
\usepackage{bbding}
\usepackage{pifont}

\newcommand{\cmark}{\ding{51}}%

\definecolor{red}{rgb}{0.863, 0.075, 0.235}
\definecolor{background}{RGB}{249, 250, 250}
\definecolor{string}{RGB}{153, 199, 180}
\definecolor{comment}{RGB}{153, 153, 153}
\definecolor{normal}{RGB}{50, 50, 50}
\definecolor{identifier}{RGB}{102, 153, 204}
\definecolor{number}{RGB}{249, 174, 87}

\newcommand{\ours}[0]{\textsc{LasTGL}\xspace}

\title{\ours: An Industrial Framework for Large-Scale Temporal Graph Learning}

\author{%
    \normalsize
    Jintang Li\textsuperscript{\textmd{1,2}}\thanks{Work done during internship research term at Ant Group}\enspace
    Jiawang Dan\textsuperscript{\textmd{2}}\enspace
    Ruofan Wu\textsuperscript{\textmd{2}}\enspace
    Jing Zhou\textsuperscript{\textmd{2}}\enspace
    Sheng Tian\textsuperscript{\textmd{2}}\enspace
    Yunfei Liu\textsuperscript{\textmd{2}}\\
    \normalsize
    \textbf{Baokun Wang}\textsuperscript{\textmd{2}}\thanks{Corresponding author}\enspace
    \textbf{Changhua Meng}\textsuperscript{\textmd{2}}\enspace
    \textbf{Weiqiang Wang}\textsuperscript{\textmd{2}}\\
    \textbf{Yuchang Zhu}\textsuperscript{\textmd{1}}\enspace
    \textbf{Liang Chen}\textsuperscript{\textmd{1}}$^\dagger$\enspace
    \textbf{Zibin Zheng}\textsuperscript{\textmd{1}}\\
    \textsuperscript{\textmd{1}} Sun Yat-sen University\\
    \textsuperscript{\textmd{2}} Ant Group\\
    \normalsize\rule{0pt}{1em}
    \faEnvelope[regular]{} Primary contact: \texttt{lijt55@mail2.sysu.edu.cn}\\
}
\begin{document}

\maketitle

\begin{abstract}
  Over the past few years, graph neural networks (GNNs) have become powerful and practical tools for learning on (\textit{static}) graph-structure data. However, many real-world applications, such as social networks and e-commerce, involve \textit{temporal} graphs where nodes and edges are dynamically evolving. Temporal graph neural networks (TGNNs) have progressively emerged as an extension of GNNs to address time-evolving graphs and have gradually become a trending topic in both academics and industry. 
  Advancing research and application in such an emerging field necessitates the development of new tools to compose TGNN models and unify their different schemes for dealing with temporal graphs.
  In this work, we introduce \ours, an industrial framework that integrates unified and extensible implementations of common temporal graph learning algorithms for various advanced tasks. 
  The purpose of \ours is to provide the essential building blocks for solving temporal graph learning tasks, focusing on the guiding principles of user-friendliness and quick prototyping on which PyTorch is based. In particular, \ours provides comprehensive temporal graph datasets, TGNN models, and utilities along with well-documented tutorials, making it suitable for both absolute beginners and expert deep learning practitioners alike.
\end{abstract}

\section{Introduction}

Graphs are a kind of data structure that models a set of objects (nodes) and their relationships (edges). Substantial endeavors have been made to learn meaningful representations to facilitate graph-based analysis tasks, with promising examples including graph neural networks (GNNs)~\cite{gcn,gat}. As a class of neural networks designed to operate directly on graph-structured data, GNNs have achieved remarkable success and have established new state-of-the-art performance across a broad spectrum of graph-based learning tasks~\cite{ogb}.

While significant progress has been made in researching \textit{static} graphs, many real-world networks, such as social, traffic, and financial networks may exhibit \textit{temporal} behaviors that carry valuable time information~\cite{EvolveGCN,survey}.
In a financial network, for example, an incoming event involves newly made payee-payor transactions in which every online transaction corresponds to a directed edge from the paying party to the receiving party in this graph~\cite{step}.
This gives rise to temporal (dynamic) graphs,
wherein the nodes and edges of the graph may undergo constant changes over time\footnote{
For the remainder of this paper, we will misuse the terminologies of ``dynamic graph'' and ``temporal graph'' interchangeably without differentiating their differences.}.
In applications where dynamic graphs arise, modeling and exploiting the temporal nature of the continuously evolving graph is crucial in representing the underlying data and achieving high predictive performance~\cite{jodie,caw}.

Owing to the widespread existence of dynamic network applications, there has been a growing interest in developing algorithms and techniques that can effectively handle Time-varying graph structures\cite{DynamicTriad,jodie,tgat,roland,spikenet}. Over the past few years, along with the success of GNNs in static graph representation learning, 
a considerable amount of research efforts have been dedicated to temporal graph representation learning.
Temporal graph learning (TGL), in general, aims to learn a compact representation that captures both the topological and dynamic information over time. In many TGL tasks, such as dynamic node classification and future link prediction, temporal graph neural networks (TGNNs), a time-oriented variant of GNNs, are the natural fit for the ever-changing temporal graphs and have achieved significantly superior performance compared to static GNNs.

Yet, research on temporal graph representation learning, particularly on TGNNs, has largely lagged behind that of their static counterparts. 
In fact, the advances in research on static graph representation learning owe much to the advent of high-quality open-source libraries/toolboxes. 
Examples include PyTorch Geometric (PyG)~\cite{pyg} and Deep Graph Library (DGL)~\cite{dgl}, which has strongly accelerated the graph learning research and industrial application by providing efficient easy-to-use building blocks for GNNs.
The growing popularity of GNNs gave researchers and developers a plethora of Python libraries to work with, which in turn facilitates graph learning research.
As a vibrant and young field, facilitating research on TGL calls for the development of domain-specific libraries that are able to deal with temporal graphs in a convenient and scalable way.

Several attempts have been made to establish a TGL benchmark or toolkit. PyTorch Geometric Temporal (PyGT)~\cite{pygt} is among the most popular libraries in dealing with discrete snapshot-based temporal graphs. DGB~\cite{DGB} and TGB~\cite{tgb} are proposed as new temporal graph benchmarks for evaluating TGL performance. Most recently, TGL~\cite{tgl} and DyGLib~\cite{dyglib} are proposed to promote reproducible research in temporal graph learning.
Despite previous works, there are still several dilemmas to be addressed in this research area.
\begin{itemize}
  \item \textbf{Lack of a well-defined abstraction for temporal graphs.}
  One of the reasons for the success of PyG and DGL, two leading graph learning libraries, is that they provide high-level abstractions for graph data.
  These abstractions streamline manipulating and analyzing (static) graph data. However, when it comes to temporal graphs, the development of similar abstractions remains largely unexplored in current libraries.
  \item \textbf{Lack of an easy-to-use toolbox with a low learning curve.}
  Switching from one toolbox to another often requires substantial effort and investment. Existing works often present steep learning curves, with a lack of user-friendly interfaces, comprehensive documentation, and accessible implementations, particularly for beginners.
  This motivates the need for a flexible and simple toolbox to facilitate widespread adoption in both academia and industry.  
  \item \textbf{Lack of comprehensive and continuous integration of various TGL methods.}
  Many existing works claim to be comprehensive but often provide only limited and outdated implementations of TGNNs for users. This points to a gap in the current research landscape, suggesting a need for more extensive and up-to-date approaches that cover a wider range of TGL methods.
\end{itemize}

\paragraph{Contributions.} In summary, our contributions are the following: (i)~\textbf{TGL cookbook}. We provide a brief summary of TGL problems and tasks, along with key milestones in recent years. This work can serve as a good starting point and a practical guide for unsophisticated users working with temporal graphs.
(ii)~\textbf{TGL toolkit.} We introduce \ours, a research- and industry-oriented library that incorporates unified and extensible collections of common TGL datasets, algorithms, and utilities for several advanced tasks. 
\ours is designed to be extensible by researchers, simple for practitioners, and fast and robust in industrial deployments.
(iii)~\textbf{Open source code.} 
We publicly release \ours with comprehensive documentation and accompanying examples for ease of use. Our code allows other researchers to reproduce our results, build on our work, and use our code to develop their own TGL models. We hope these works together can encourage more researchers to investigate in this direction.
(iv)~\textbf{TGL ecosystem.} 
\ours is more than just a toolkit; it is a growing ecosystem that aims to bridge the gap between cutting-edge research and practical applications in temporal graph learning. Through continuous updates and community engagement, we envision \ours becoming an indispensable resource for both academics and industry professionals working in this dynamic field.

The remainder of the paper is structured as follows. Necessary background knowledge and related work are presented in Section~\ref{sec:preliminary} and Section~\ref{sec:related_work}, respectively. In Section~\ref{sec:architecture}, \ours is detailed and accompanied by code snippets for illustration. Finally, we draw our conclusion in Section~\ref{sec:conclusion}.

\section{Preliminaries}
\label{sec:preliminary}
In this section, we present concepts, notations, and problem settings used in the work. 
We primarily delve into the (dynamic) node classification and future link prediction tasks on the temporal graphs, which are among the most important downstream tasks in the temporal graph learning domain.

\subsection{Temporal graphs}
Temporal graphs are an effective data structure to represent the evolving topology/relationships between various entities across time dimensions.
Different applications give rise to different types of temporal graphs.
As pointed out in~\cite{KazemiGJKSFP20,survey}, reasoning over temporal graphs typically falls under two settings: discrete-time and continuous-time settings, which represent a temporal graph as a sequence of discrete snapshots and an asynchronous stream of timed events, respectively.

\begin{definition}[Discrete-time temporal graph (DTTG)]
  A temporal graph over $T$ times can be denoted as an ordered sequence of static graph snapshots observed at discrete time points., i.e., $\mathcal{G}=\{\mathcal{G}_1, \mathcal{G}_2,\ldots\}$, where each $\mathcal{G}_t=(\mathcal{V},\mathcal{E}_t)$ is a (static) graph, referred as a \textit{snapshot}. For simplicity, $\mathcal{G}_t \in \mathcal{G}$ is defined over the same set of $N$ vertices $\mathcal{V}$, but may differ in terms of the graph topology $\mathcal{E}_t$.
\end{definition}

\begin{definition}[Continuous-time temporal graph (CTTG)]
  A temporal graph over $T$ times can alternatively be denoted as a stream of dyadic events happening sequentially, i.e., $\mathcal{G}=\{(e_1, t_1), (e_2, t_2),\ldots\}$. Each $e_t$ is a temporal event represented as a tuple of node(s) and associated node/edge observation type, e.g., $(u, $`add'$)$, $(u, v, $`add'$)$, or $(u, v, $`delete'$)$. This allows for a fine-grained representation of temporal information, capturing events and changes at precise points in time. 
\end{definition}

\paragraph{Comparison between DTTGs and CTTGs}
Both DTTGs and CTTGs have their own unique challenges and require different analysis techniques.
The major difference between them is that the edges and nodes in a CTTG are associated with timestamps, representing a continuous flow of time, while in a DTTG, the edges and nodes are associated with discrete time intervals. Although DTTG bridges the gap between temporal and static graphs, some time-continuous information is inevitably lost due to discrete processing. By contrast,
CTTG addresses this defect in DTTG by representing the temporal graph as a time-continuous event stream and usually can achieve better results. 
In this regard, CTTGs are more prevalent for time-critical applications due to their capability of capturing more intricate and fine-grained temporal patterns. 
Generally, the DTTG can be considered as a specific case of a CTTG where observations or events occur at discrete points in time.
A graph snapshot (corresponding to a static graph) at any time point $t$ can be obtained from a continuous (streamed) temporal graph by accumulating evolution across a window of consecutive events that occurred before (or at) time $t$. In some cases, multiple edges may have been added between two nodes giving rise to multi-graphs. How to properly deal with multi-edges becomes a critical challenge.

\subsection{Types of evolution}
\begin{figure}[h]
  \centering
    \includegraphics[width=\linewidth]{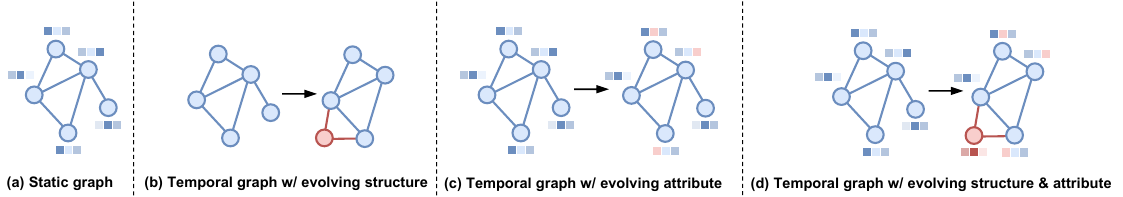}
  \caption{Comparison of (a) static graph and (b)-(d) temporal graphs with different evolving patterns.}
  \label{fig:evolution}
\end{figure}

For both DTTGs and CTTGs, different components, such as graph structure and node attributes, may undergo changes and evolve over time. 
To provide a better understanding of temporal graphs, we categorize three main types of the evolution of temporal graphs in Figure~\ref{fig:evolution} and detail them as follows.

\paragraph{Temporal graph with evolving structure}
As shown in Figure~\hyperref[fig:evolution]{1(b)}, new nodes may emerge and form connections with existing nodes, while existing nodes can disappear, resulting in the termination of their associated edges. For instance, in a social network, when users sign up for the platform and establish connections with others, it results in the creation of new nodes and edges in the graph, respectively. Conversely, when users delete their accounts, nodes, and corresponding edges are removed from the graph.

\paragraph{Temporal graph with evolving attribute}
A vast majority of real-world graphs are coupled with a rich set of attribute information, which may also evolve over time~\cite{star}. Node attributes can include user profiles, interests, geographical locations, and more, while edge attributes can describe relationship types between users and interaction frequencies. As shown in Figure~\hyperref[fig:evolution]{1(c)}, node attributes are changed although the graph topology remains fixed over a period of time.

\paragraph{Temporal graph with evolving structure \& attribute}
Typically, real temporal graphs often exhibit complex co-evolution patterns. As illustrated in Figure~\hyperref[fig:evolution]{1(d)}, the evolution occurs in both the graph topology and node attributes. 
For example, users in a social network can update their profiles and establish friendships with others simultaneously.
Consequently, the task becomes more challenging as a model has to jointly learn the node representations considering both spatial and temporal aspects.

\subsection{Temporal graph learning tasks}

\begin{figure}[h]
  \centering
    \includegraphics[width=\linewidth]{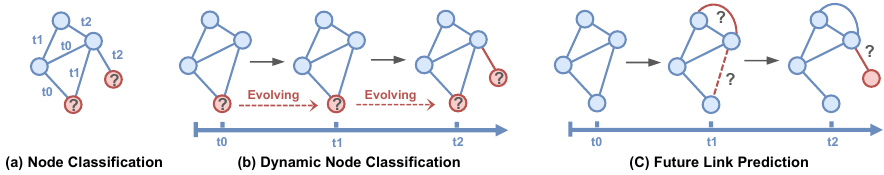}
  \caption{Illustrative examples of (a) node classification, (b) dynamic node classification, and (c) future link prediction.}
  \label{fig:task}
\end{figure}

Learning over graphs involves two fundamental tasks, including node prediction and link prediction\footnote{In this work, we consider only node-level and edge-level tasks, as graph-level tasks can be regarded as a special case of node-level tasks.}, whose objectives are to infer the properties of nodes (e.g., class labels) and edges (e.g., existence), respectively. 
In the context of temporal graphs, there are three classical learning tasks, i.e., node classification, dynamic node classification, and future link prediction. 
We provide illustrative examples of the three tasks in Figure~\ref{fig:task}.

\begin{definition}[Node classification]
For a given temporal graph $\mathcal{G}$, the goal of node classification is to classify the labels of nodes $\mathcal{V}_\text{sub} \subseteq \mathcal{V}$ based on a subset of nodes with known class labels.
\end{definition}

The node classification task in temporal graphs is a straightforward extension of static graphs, which additionally incorporates time information to assist in the classification task. In this task, the labels of nodes are fixed across a wide range of time intervals.

\begin{definition}[Dynamic node classification]
For a given temporal graph $\mathcal{G}_t$, where $\mathcal{G}_t$ can be a graph with events at timestamp $t$ in a CTTG or the $t$-th snapshot in a DTTG, the goal of dynamic node classification is to classify the labels of nodes $\mathcal{V}_t \subseteq \mathcal{V}$ based on a subset of nodes with known class labels at timestamp/snapshot $t$. 
\end{definition}

Compared to the traditional node classification task, an arguably less complex task, dynamic node classification is more challenging in terms of capturing temporal dependencies and predicting the evolving nature of nodes in a graph. It involves predicting the labels of nodes that may include previously unseen nodes or nodes with constantly changing class labels. This dynamic aspect necessitates that TGNNs adapt to the evolving graph structure and node properties to make accurate predictions in a changing environment.

\begin{definition}[Future link prediction]
For two nodes $u,v \in \mathcal{V}$ at a given timestamp $t$, future link prediction aims to predict the existence of an edge $(u,v)$ based on historical observations from all nodes and their associated links before timestamp $t$.
\end{definition}

Typically, an edge can appear or disappear multiple times at different timestamps, resulting in temporal graphs with multiple occurrences of the same edge. Handling these multiple edges requires careful consideration and specialized techniques to preserve the temporal dynamics.
The future link prediction problem is commonly tackled using CTTGs because the appearance or disappearance of links in real-world networks often happens constantly and continuously over time. In such a task, CTTGs provide a more fine-grained representation of the temporal dynamics of graph topology and enable a more accurate and comprehensive way to predict future links in dynamic networks.

\begin{figure}[t]
  \centering
  \begin{picture}(400,160)
  \put(0,0){\includegraphics[width=\linewidth]{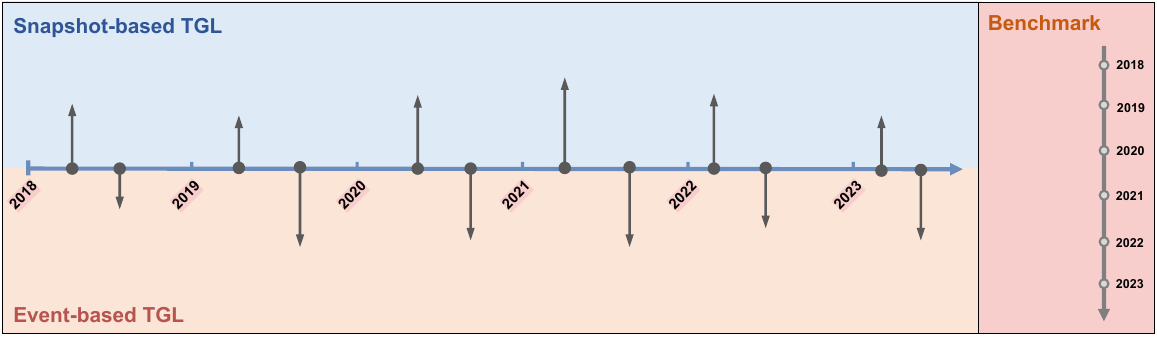}}
  \put(7,82){\scriptsize {DynamicTriad~\cite{DynamicTriad}}}

  \put(28,37){\scriptsize {HTNE~\cite{htne}}}
  \put(28,30){\scriptsize {CTDNE~\cite{CTDNE}}}
  \put(28,23){\scriptsize {NetWalk~\cite{NetWalk}}}  

  \put(65,84){\scriptsize {tNodeEmbed~\cite{tNodeEmbed}}}
  \put(72,77){\scriptsize {STAR~\cite{star}}}

  \put(90,24){\scriptsize {JODIE~\cite{jodie}}}
  \put(90,17){\scriptsize {M$^2$DNE~\cite{mmdne}}}
  \put(90,10){\scriptsize {DyRep~\cite{dyrep}}}  

  \put(125,90){\scriptsize {EvolveGCN~\cite{EvolveGCN}}}
  \put(129,83){\scriptsize {DySAT~\cite{DySAT}}}
  
  \put(150,27){\scriptsize {TGAT~\cite{tgat}}}
  \put(150,20){\scriptsize {TGN~\cite{tgn}}}
  \put(146,13){\scriptsize {dyngraph2vec~\cite{dyngraph2vec}}}  

  \put(181,97){\scriptsize {DyHINE~\cite{DyHINE}}} 
  \put(184,90){\scriptsize {TRRN~\cite{TRRN}}}   

  \put(205,24){\scriptsize {DDGCL~\cite{ddgcl}}} 
  \put(205,17){\scriptsize {CAW-N~\cite{caw}}}   
  \put(205,10){\scriptsize {APAN~\cite{APAN}}}    
  
  \put(232,91){\scriptsize {ROLAND~\cite{roland}}}
  \put(232,84){\scriptsize {CTGCN~\cite{CTGCN}}}   
  
  \put(250,31){\scriptsize {FreeGEM~\cite{freegem}}}   
  \put(250,24){\scriptsize {EdgeBank~\cite{DGB}}}   
  \put(250,17){\scriptsize {PINT~\cite{PINT}}}   
  
  \put(288,97){\scriptsize {TiaRa~\cite{TiaRa}}}    
  \put(288,90){\scriptsize {WinGNN~\cite{WinGNN}}}    
  \put(288,83){\scriptsize {SpikeNet~\cite{spikenet}}}
  \put(288,76){\scriptsize {DyTed~\cite{DyTed}}}   
  
  \put(305,25){\scriptsize {SAD~\cite{sad}}}  
  \put(302,18){\scriptsize {TGSL~\cite{tgsl}}} 
  \put(300,11){\scriptsize {TIGER~\cite{TIGER}}}

  \put(342,93){\scriptsize {DyGEM~\cite{DynamicGEM}}} 
  
  \put(347,80){\scriptsize {PyG~\cite{pyg}}}  
  \put(347,73){\scriptsize {DGL~\cite{dgl}}} 
  
  \put(347,62){\scriptsize {OGB~\cite{ogb}}}      
  \put(344,48){\scriptsize {PyGT~\cite{pygt}}}
  
  \put(347,36){\scriptsize {DGB~\cite{DGB}}}  
  \put(347,29){\scriptsize {TGL~\cite{tgl}}}    
  \put(347,17){\scriptsize {TGB~\cite{tgb}}}    
  \put(343,10){\scriptsize {DyGLib~\cite{dyglib}}}    
  \put(345,3){\scriptsize \color{red}\textbf{\underline{\ours}}}   
  \end{picture}  
  \caption{Milestones of TGL models and benchmarks in the last few years. For research articles that were initially available as preprints, such as those on \href{https://arxiv.org/}{arXiv}, we record their publication year based on when they appeared in \textit{conferences} or \textit{journals} in this timeline. (DyGEM: DynamicGEM; PyGT: PyTorch Geometric Temporal)}
  \label{fig:timeline}
\end{figure}

\section{Related work}
\label{sec:related_work}

As research on temporal graphs has become increasingly prevalent, temporal graph learning has seen rapid development in recent years.
In what follows, we briefly review cutting-edge research on temporal graph representation learning, as well as graph neural network libraries and benchmarks.

\subsection{Temporal graph representation learning}
We first provide a literature overview of several well-known approaches proposed in the literature to present major progress in TGL. Following the taxonomy introduced in \cite{survey}, mainstream TGL methods can be roughly categorized into two broad classes, including discrete-time (snapshot-based) and continuous-time (event-based) methods. 
Milestones in TGL models and benchmarks over the last few years are illustrated in Figure~\ref{fig:timeline}.
We try to summarize the most notable ones without any claim that this list is complete.
Note that this paper is a snapshot of the state-of-the-art of TGL at the time of writing and will almost certainly be outdated soon. 
We kindly refer readers to \cite{DBLP:journals/corr/abs-2302-01018} and \cite{survey} for comprehensive surveys of TGL research.

\paragraph{Snapshot-based TGNNs}
Snapshot-based TGNNs are used to learn over sequences of time-stamped static graphs (i.e., DTTGs), with a mechanism that learns the temporal dependencies across different time steps. Snapshot-based TGNNs have been extensively studied in the literature due to their flexibility in modeling temporal graphs and ease of implementation~\cite{spikenet}.
Early works mainly focus on learning node representations by simulating temporal random walks~\cite{tNodeEmbed} or modeling the triadic closure process~\cite{DynamicTriad} on multiple graph snapshots.
These methods typically generate piecewise constant representations and may suffer from the staleness problem~\cite{survey}.
In recent years, the most established solution has been switched to combine sequence models (mostly RNNs~\cite{cho2014learning}) with static GNNs to capture temporal dependencies and correlations between snapshots~\cite{star,EvolveGCN,CTGCN}. For example, EvolveGCN~\cite{EvolveGCN} extends graph convolution networks (GCN) to temporal graphs with RNNs learning the evolving parameters. In addition, self-attention is also employed as a building block for dealing with sequential relationships between snapshots~\cite{DySAT,TRRN,DyHINE}.
As RNN-based sequence models require high computational and memory overheads for sharing temporal contextual information across snapshots, SpikeNet~\cite{spikenet} instead adopts spiking neural networks to efficiently capture the evolving dynamics underlying a discrete graph sequence. 
Despite the preliminary success, current works on DTTGs have limitations in both their design and training strategies without translating fully the success achieved on static graphs~\cite{roland}. In this regard, recent frameworks such as ROLAND~\cite{roland} and its variants~\cite{WinGNN,CS-TGN} have been proposed to repurpose static GNNs to temporal graphs.

\paragraph{Event-based TGNNs}
Another major branch of TGL is event-based methods, which process the time-evolving graphs as sequences of streaming events (i.e., CTTGs) and update a node or/and an edge if an event occurs involving that node or edge.
To incorporate the temporal information, temporal random walks are sampled to serve as the contextual information of nodes~\cite{CTDNE,NetWalk,caw} to learn time-dependent embeddings. 
Owing to their strong capability in capturing fine-grained temporal dynamics, research on CTTGs has been dominated by RNN-based sequence approaches as well~\cite{jodie,tgat,tgn}.
JODIE~\cite{jodie} pioneers to leverage two RNNs to sequentially update the node embeddings at every bipartite interaction. TGAT~\cite{tgat} extends the graph attention mechanism~\cite{gat,GAAN} to temporal graphs with a functional time encoding technique to encode time in an embedding space. As a generalization of JODIE and TGAT, TGN~\cite{tgn} then proceeds by introducing a memory module that represents the state of the node at a given time, acting as a compressed representation of the node's historical interactions.
Another line of research in this area is by modeling the event sequence as a temporal point process (TPP)~\cite{htne,mmdne,dyrep}. Typically, a TPP is a stochastic process~\cite{Guttorp90} composed of a time series of binary events that occur in continuous time, aiming to capture the interleaved dynamics of the observed processes by regarding the arrival of nodes/edges as an event.
HTNE~\cite{htne} model nonlinearly evolving dynamic processes and learns to encode structural-temporal information over temporal graphs via a Hawkes process.
DyRep~\cite{dyrep} and M$^2$DNE~\cite{mmdne} propose to capture structural and temporal evolutionary patterns of graphs with a temporal attention point process model.
Spurred by the rise of self-supervised learning on graphs, there are also works exploring graph contrastive learning on temporal graphs~\cite{ddgcl,sad,step}.

\subsection{Graph neural network libraries and benchmarks}
Fundamental breakthroughs in graph learning research have been facilitated by the availability of many high-quality open-source libraries and graph benchmarks. 
Over the last several years, a bunch of Python libraries for deep graph learning have emerged for either TensorFlow~\cite{tensorflow} or PyTorch~\cite{pytorch} to work with graphs, leading to rapid growth in the field of (static) graph representation learning~\cite{pyg,dgl}.
The most popular open-source libraries for GNNs at present are PyTorch Geometric (PyG)~\cite{pyg} and Deep Graph Learning Library (DGL)~\cite{dgl}, both of which enable researchers and practitioners to easily work with graph-structured data with a rich set of functionalities and pre-built modules.
In parallel with the success of PyG and DGL, several well-curated datasets, such as Open Graph Benchmark (OGB)~\cite{ogb}, Long Range Graph Benchmark (LRGB)~\cite{lrgb}, and IGB~\cite{igb}, have been proposed to further accelerate the advancements in deep graph learning.

Yet, research on temporal (dynamic) graph learning has largely lagged behind its static counterparts. 
Although PyG and OGB have some support for TGNNs and temporal datasets, there is still a need for a comprehensive framework and a dedicated benchmark specifically tailored for temporal graphs. 
The growing popularity of TGNNs necessitates the availability of a specialized software library to effectively work with them.
DynamicGEM~\cite{DynamicGEM} is one of the pioneering frameworks for dynamic network representation learning, which extends graph embedding methods to handle the evolution of temporal graphs. PyTorch Geometric Temporal (PyGT)~\cite{pygt} is an extension of PyG for spatiotemporal signal processing, with a main focus on implementing basic operations for building snapshot-based TGNNs. Both frameworks have primarily been developed and optimized for small-scale graphs, which limits their applicability in real-world scenarios where large-scale temporal graphs are prevalent.
To this end, TGL~\cite{tgl} has been further proposed as a general framework for offline TGNN training, which offers several efficient operations specifically designed for training on large-scale temporal graphs. 
However, the codebase in TGL currently supports only event-based TGNNs, such as TGN and TGAT, which are not considered state-of-the-art in this field of research.

Benchmark datasets for temporal graphs have also emerged with the development of TGL frameworks. A recent work~\cite{DGB} first identified challenges and drawbacks in the current evaluation setting for dynamic link prediction and introduced a Dynamic Graph Benchmark from a diverse set of domains. As DGB focuses on link-level tasks and provides datasets on a small scale, Temporal Graph Benchmark (TGB)~\cite{tgb} was introduced to continue the success of large-scale benchmarks (e.g., OGB) in temporal graph learning research. TGB establishes a comprehensive and diverse million-scale graph benchmark that enables standardized evaluation and comparison of various methods, providing new challenges and opportunities for future research. 
As benchmarks continue to expand in size, it is necessary to develop new scalable TGL library to effectively handle large-scale temporal graphs.

\begin{figure}[t]
  \centering
    \includegraphics[width=0.83\linewidth]{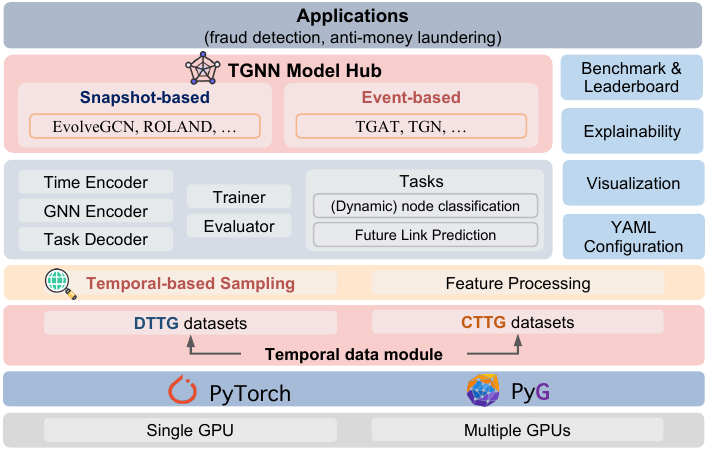}
  \caption{Overview of \ours framework.}
  \label{fig:framework}
\end{figure}

\section{\ours: design and architecture}
\label{sec:architecture}

In this section, we aim first to provide an overview of the \ours library including the overall design and architecture. As illustrated in Figure~\ref{fig:framework}, our \ours is built upon PyTorch and PyG, both of which provide highly efficient API for implementing (graph) neural network architectures. 
\ours is rigorously tested and thoroughly documented.
Most of the APIs of \ours are designed to be compatible with PyG and PyTorch to lower the learning curve of users from both communities and leverage the rich ecosystem of tools and resources. 
Particularly, PyG users can seamlessly experiment with \ours in temporal graph scenarios with minimal effort.

We present the main characteristics of \ours as follows: (i) a collection of temporal graph datasets that include both CTTGs and DTTGs benchmarks and an advanced temporal data module to work with (Section~\ref{sec:dataset} and \ref{sec:data}); (ii) Neighborhood sampling and feature processing techniques for tackling the structure and feature of temporal graphs (Section~\ref{sec:sampling} and \ref{sec:feature}); (iii) TGNN model hub including state-of-the-arts; (iv) Scalable and fully reusable training and inference pipelines with configurable options (Section~\ref{sec:model} and \ref{sec:pipeline});  (v) Explanation \& visualization facilities (Section~\ref{sec:explain} and \ref{sec:vis}).

\subsection{Temporal graph datasets}
\label{sec:dataset}
Many sophisticated TGL libraries have integrated their own temporal graph datasets, which serve as standardized benchmarks for the inference of the performance and effectiveness of TGL models. PyG and DGL also offer partial support for working with temporal datasets. However, the majority of existing datasets (e.g., DGB and TGB) are formulated as CTTGs, which poses challenges when applying snapshot-based TGNNs that are naturally defined on DTTGs. We note that PyGT has released several DTTG-based datasets, which are relatively small-scale and only support node regression tasks.

To enable a more comprehensive and reliable evaluation of TGL models, \ours has integrated a diverse set of temporal graph benchmark datasets. This includes several large-scale DTTG benchmark datasets as well as many popular CTTG datasets. \ours is a highly compatible and extendable framework that seamlessly integrates with existing dataset collections and can be easily extended, which enables the platform to remain up-to-date with the evolving needs and challenges of the whole TGL community.

\subsection{Advanced temporal data module}
\label{sec:data}
Graphs are complex data objects that require advanced data modules to accurately describe, access, and manipulate their internal structures. 
For example, PyG offers its own specialized data modules, namely \texttt{Data} and \texttt{HeteroData}, designed to handle standard (homogeneous) and heterogeneous graphs, respectively. Similarly, DGL has also provided its own \texttt{DGLGraph} data module for storing graph information.
These data modules provide the necessary functionality and utilities to effectively work with different types of graphs, ensuring efficient operations and computations. 

In the case of temporal graphs, which involve time-dependent changes and evolving dynamics, a suitable data module becomes even more essential. It should be capable of capturing and representing temporal aspects, such as time-stamped edges, evolving node attributes, and dynamic graph structures.
In \ours, we introduce the \texttt{TemporalData} data module following the design philosophy of PyG to tackle temporal graphs. 
\texttt{TemporalData} enables easy conversion between CTTGs and DTTGs. Users can easily transform a CTTG into a DTTG representation by discretizing the timestamps, or vice versa by assigning continuous timestamps to a DTTG. With \texttt{TemporalData}, users can also convert a temporal graph, whether it's a CTTG or DTTG, to its static counterpart by simply discarding the temporal dimension. We provide a pseudocode example in Figure~\ref{fig:temporal_data}.
\begin{figure}[h]
\centering
\begin{lstlisting}
from lastgl.data import TemporalData

data = TemporalData(src=..., dst=..., t=...) # CTTGs-represented data
static_data = data.to_static() # To PyG static data
snapshots = data.snapshots(time_attr='t') # To DTTGs
\end{lstlisting}
\vspace{-8mm}
\caption{Main characteristic of \texttt{TemporalData}.}
\label{fig:temporal_data}
\end{figure}

By leveraging \texttt{TemporalData}, 
researchers and practitioners are provided with a powerful tool to effectively model and analyze temporal dependencies in graph data.
The development of a dedicated data module specifically designed for temporal graphs would enable seamless handling of time-based information, facilitating tasks such as dynamic node classification and future link prediction.

\subsection{Temporal-based sampling}
\label{sec:sampling}
\begin{wrapfigure}{r}{0.3\linewidth}
    \centering
    \vspace{-5mm}
    \includegraphics[width=\linewidth]{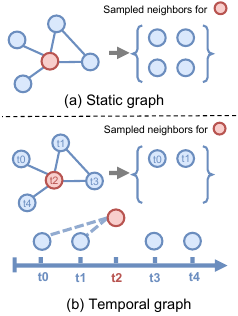}
    \caption{Neighborhood sampling on (a) static graph and (b) temporal graph, respectively. }
    \vspace{-12mm}
    \label{fig:sampling}
\end{wrapfigure}

\subsubsection{Temporal neighborhood sampling}
Neighborhood sampling is a fundamental technique in modern GNNs for tackling the scalability issue, which allows for mini-batch training of GNNs on large-scale graphs where full-batch training is not feasible. For each iteration, a subgraph with a fixed-size set of neighbors is sampled for computing, alleviating the need to operate on the entire graph during training. However, it is non-trivial to perform neighborhood sampling on temporal graphs as the time information of the neighbors needs to be considered. Specifically, nodes are capable of leveraging neighborhood information up to the present moment when predicting future events.
As shown in Figure~\ref{fig:sampling}, sampling on temporal graphs requires additional constraints where only neighbors in the past can be sampled to avoid data leakage from future information.

To address this problem, \ours incorporates functional temporal neighborhood sampling, which takes into account both the structural connectivity and the temporal proximity during the sampling process. 
This ensures that the sampled neighbors exhibit strong connections with the reference node and align with the desired temporal patterns.
Specifically, we design two fundamental sampling strategies for temporal graphs, \textit{uniform sampling} where neighbors in the past are sampled uniformly given a time window, and \textit{most-recent sampling} where only the most recent neighbors are sampled.
Figure~\ref{fig:sampling_code} compares the neighborhood sampling process of \ours and PyG. \ours closely follows the design philosophy and data structure of PyG, but additionally provides the support of neighborhood sampling on temporal graphs. 
PyG users can take full advantage of \ours's capabilities in temporal graph scenarios by only modifying very few lines of code.
To this end, \ours enables TGNNs to better capture and analyze time-dependent phenomena of temporal graphs.

\begin{figure}[h]
\centering
\begin{subfigure}[t]{0.54\textwidth}
\centering
\begin{lstlisting}
loader = TemporalLoader(temporal_data, [10, 10])
for batch in loader:
    batch = batch.to(device)
    output = model(batch.x, 
                   batch.edge_index, 
                   batch.t) # time information
\end{lstlisting}
\vspace{-5mm}
\caption{\ours}
\end{subfigure}
\hspace{1mm}
\begin{subfigure}[t]{0.42\textwidth}
\centering
\begin{lstlisting}
loader = NeighborLoader(data, [10, 10])
for batch in loader:
    batch = batch.to(device)
    output = model(batch.x, 
                   batch.edge_index,
                   )
\end{lstlisting}
\vspace{-5mm}
\caption{PyG}
\end{subfigure}
\caption{Neighborhood sampling in (a) \ours and (b) PyG for temporal and static graphs, respectively.}
\label{fig:sampling_code}
\end{figure}

\subsubsection{Temporal negative sampling}
Negative sampling is a method commonly used to \textit{train} and \textit{evaluate} TGL methods, which is as important as positive sampling in graph representation learning~\cite{YangDZYZT20}. 
In the context of temporal graphs, where the structure of the graph evolves over time, negative sampling involves selecting edges that do not exist at the current time point but could potentially have existed. Negative samples are typically incorporated into the training objective function alongside positive samples to provide the TGL model with the ability to distinguish between positive (actual) and negative (non-existent) connections in the evolving graph. In the evaluation phase, it is also crucial to include appropriate negative samples for a rigorous and robust assessment of TGL models~\cite{DGB,tgb}. 

\begin{figure}[h]
  \centering
    \includegraphics[width=0.9\linewidth]{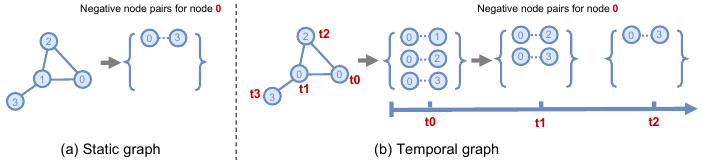}
  \caption{Negative sampling on (a) static graph and (b) temporal graph, respectively.}
  \label{fig:negative_sampling}
\end{figure}

However, implementing negative sampling on temporal graphs is non-trivial due to the temporal constraints. As illustrated in Figure~\ref{fig:negative_sampling}, unlike static graphs where negative edges can be randomly sampled, the negative samples for a node dynamically change along with the evolving graph structure. This introduces additional challenges in the design and implementation of effective negative sampling strategies for temporal graphs.

In light of the current success in TGL~\cite{tgb,DGB}, \ours provides two different sampling strategies for incorporating temporally non-existent edges: (i) \textit{random negative sampling}, which uniformly samples edges absent before the current time step; (ii) \textit{historical negative sampling}, which samples historical edges observed during previous timestamps but are absent in the current step. 
Both strategies can be conveniently employed to assist in adapting the TGL model to temporal changes during training or to measure its capacity to understand the evolving graph structure during evaluation.

\subsection{Feature preprocessing}
\label{sec:feature}
Temporal graphs often involve a complicated set of engineering problems when dealing with node, edge, and graph features. Typically, raw features in industrial scenarios are represented using various semantic types, including numerical (e.g., price), categorical (e.g., gender or occupation), and textual (e.g., descriptions) attributes. However, modern neural networks tend to achieve better performance when applied to numerical inputs.
Even on public graph benchmarks that contain numerical features, employing dedicated feature preprocessing techniques such as normalization can result in state-of-the-art performance with a simple model~\cite{DBLP:journals/corr/abs-2309-00367}.
Feature engineering is one necessary step to improve the performance of downstream TGNNs.
In this regard, \ours incorporates certain functionalities to build a TGL framework to perform effective machine learning on such complex data.

\subsection{TGNNs model hub}
\label{sec:model}
To make TGL research scalable and fast to iterate on, \ours provides a large set of modularized TGNNs implementations including event-based and snapshot-based ones. 
All TGNNs closely follow the official implementation with the least effort for adapting the architecture to our codebase. Typically, we divide a TGNN into three main components: time encoder, graph encoder, and task decoder.
Wherever possible, different architectures follow the same API allowing users to switch easily between different models. 
To maintain flexibility and extensibility, all the implemented TGNN models are purely built upon \texttt{torch.nn.Module} module. Users can directly apply TGNN models in our model hub to their own frameworks as long as PyTorch is the same supported backend.
Note that more interesting directions and algorithms can be easily incorporated into \ours based on the unified and extensible implementations.
Our overall framework is easily extendable to new TGNN models where new components can be created and combined with already existing ones.

\subsection{Training and inference pipelines}
\label{sec:pipeline}
Although PyG is extremely easy to use to build complex GNN variants, it does not provide training and inference workflows for developers working on different tasks. It is tedious and time-consuming as developers have to build their own pipelines from scratch. 
It remains a major inconvenience for researchers or developers to implement algorithms and empirically compare to baselines, and for beginners to quickly get started learning TGNNs. 
To meet the needs of both beginners and advanced users, \ours provides a fully reusable and standardized pipeline by structuring the code to abstract the details of training and inference:
\begin{figure}[h]
\centering
\begin{lstlisting}[language=Python,breaklines=True]
from lastgl.config import load_config
from lastgl.training import Trainer
from lastgl.models.event_based import TGAT

cfg = load_config('path_to_yaml_file')
model = TGAT(cfg)
trainer = Trainer(cfg)
trainer.fit(train_data)
trainer.evaluate(test_data)
\end{lstlisting}
\vspace{-8mm}
\caption{An easy-to-use pipeline through \texttt{Trainer} with configurations.}
\label{fig:pipeline}
\end{figure}

To further enhance the reproducibility and scalability of experiments and facilitate the development of state-of-the-art TGNNs, \ours offers the flexibility to customize the pipeline according to specific requirements through desired configurations. Specifically, users can easily configure \ours to train and tune different TGNN variants by YAML configuration files. One can also conduct experiments across various tasks, such as dynamic node classification and future link prediction, by easily overriding the configuration files.

\subsection{Explainable TGNNs}
\label{sec:explain}
Neural networks, including TGNNs, are often criticized due to their inherent complexity and black-box nature. 
Specifically, the predictions made by these models are often hard to interpret. 
This has led to a growing demand for better interpretability and explainability to understand how these models make decisions or reason, particularly in domains such as healthcare, finance, or self-driving vehicles where reliable decision-making is critical. 
However, while many efforts have been made to explain the prediction mechanisms of (static) GNNs, there has been relatively little work done for TGNNs.
In this regard, \ours integrates a suite of explainability algorithms to provide holistic explanations for TGNNs. The provided algorithms are extensions of those for static ones (e.g., GNNExplainer~\cite{gnnexplainer}), which additionally consider the temporal dimension and update schemes.
They can be applied to broad TGNN models of any type and used both in research and production environments, which paves the way for building interpretable TGNNs and establishing trustworthy graph-based systems.

\subsection{Visualization of data}
\label{sec:vis}
Visualization of graph data is an important problem in various fields, including bioinformatics, software engineering, database and web design, machine learning, and visual interfaces for other technical domains.
As explanations play a pivotal role in understanding the behavior of TGNNs, visualization can also assist in explaining the nodes/edges that have the most significant contribution towards the prediction.
In practice, it would be much more convenient for users to visualize and analyze their (temporal) graphs effectively.
However, unlike image data, which can be intuitively visualized, visualizing graph data has been so far a challenge. In graph data, the relationships between nodes and edges are more complex and require specialized techniques for effectively visualizing the graph layout. 
To further facilitate the progress of TGL research, \ours, offers specifically designed APIs that address the challenges associated with visualizing graph data in an intuitive and interpretable manner.

\subsection{Comparison to other libraries}
\label{sec:comparison}
Given the growing popularity of the field, several libraries for TGL have appeared in recent years to ease the implementations of TGNNs.
Probably the closest work to \ours is PyGT, which also focuses on simplifying research with several flexible operations defined on temporal graphs. 
In addition, there are other related approaches to \ours, such as TGL and DynamicGEM, which also provide event-based TGNN implementations. However, many of these methods do not specifically differentiate between continuous and discrete temporal graphs. Instead, they primarily focus on a single type of TGNN with limited scalability and functionalities, thus restricting their applicability in various scenarios. 
More importantly, to the best of our knowledge, there has been limited research attention given to industrial scenarios that are fraught with unique challenges. These challenges include handling giant-scale graphs, managing complex and heterogeneous data, and enabling explainable and reliable decision-making processes. 
We summarize the characteristics of mainstream toolkits and libraries in Table~\ref{tab:comparison}. Compared to existing libraries, \ours offers comprehensive support for both snapshot- and event-based TGNNs. Furthermore, \ours is a scalable and extensible benchmark that incorporates several functionalities, including a unified training/inference pipeline, visualization, and explanation tools. 
With \ours, it is effortless for researchers and engineers to benchmark TGNNs, and also for beginners to quickly learn to go deep into this field.

\begin{table*}[h]
    \centering
    \caption{Comparison of existing TGNN libraries and \ours.}\label{tab:comparison}
    \resizebox{\linewidth}{!}
{\begin{tabular}{lccccccc}
\toprule
            & \textbf{Snapshot-based} & \textbf{Event-based} & \textbf{Unified pipeline} & \textbf{Benchmark} & \textbf{Scalability} & \textbf{Visualization} & \textbf{Explainability} \\
\midrule
PyGT~\cite{pygt}       &    \cmark                 &     -             &      -           &      \cmark     &     -        &     -          &    -            \\
DynamicGEM~\cite{DynamicGEM} &          -           &    \cmark             &        -         &    -       &     -        &     -          &    -            \\
TGL~\cite{tgl}        &    -                 &     \cmark             &       -          &     -      &      \cmark        &       -        &     -           \\
DyGLib~\cite{tgl}        &    -                 &     \cmark             &       -          &     \cmark      &              &       -        &     -           \\
\midrule
\ours      &  \cmark                   &    \cmark              &     \cmark            &     \cmark      &       \cmark      &    \cmark           &      \cmark          \\ 
\bottomrule
\end{tabular}
    }

\end{table*}

\section{Conclusion}
\label{sec:conclusion}
In this paper, we present a concise overview of temporal graph learning problems and tasks, as well as the key milestones achieved in recent years. We also introduce \ours, an open-source toolkit tailored for working with temporal graphs. \ours implements a wide range of methods for temporal graph representation learning, including both snapshot-based and event-based TGNNs, and provides utilities for processing temporal graphs as well as loading popular benchmark datasets. With a comprehensive model hub and a corresponding training and evaluation pipeline, we aim to accelerate temporal graph learning workflows in a flexible and user-friendly manner. In production, users can effortlessly reuse, apply, and scale these implementations to real-world applications with larger graphs. Furthermore, \ours is designed as a dynamic framework that will be regularly updated with new metrics, datasets, and models to keep pace with the evolving research in this field.

\bibliographystyle{abbrv}
\bibliography{main}

\end{document}